\documentclass[lettersize,journal]{IEEEtran}
 \UseRawInputEncoding

\usepackage{cite}
\usepackage{amsmath,amssymb,amsfonts}
\usepackage{mathtools}
\usepackage{algorithmic}
\usepackage{algorithm}
\usepackage{array}
\usepackage[caption=false,font=normalsize,labelfont=sf,textfont=sf]{subfig}
\usepackage{textcomp}
\usepackage{stfloats}
\usepackage{url}
\usepackage{rotating}
\usepackage{adjustbox}
\usepackage{xcolor}
\usepackage{booktabs}
\usepackage{ wasysym }
\usepackage[flushleft]{threeparttable}
\usepackage{makecell}
\usepackage{tabularx}
\usepackage{bm,times} 
\usepackage{ upgreek }
\usepackage{multirow}
\usepackage{verbatim}
\usepackage{multicol}
\usepackage{wrapfig}
\usepackage{relsize}
\usepackage{graphicx}
\usepackage[short]{optidef}
\usepackage{hyperref}

\RequirePackage[normalem]{ulem} 
\RequirePackage{color}\definecolor{RED}{rgb}{1,0,0}\definecolor{BLUE}{rgb}{0,0,1} 
\providecommand{\DIFaddbegin}{} 
\providecommand{\DIFaddend}{} 
\providecommand{\DIFdelbegin}{} 
\providecommand{\DIFdelend}{} 
\providecommand{\DIFaddbeginFL}{} 
\providecommand{\DIFaddendFL}{} 
\providecommand{\DIFdelbeginFL}{} 
\providecommand{\DIFdelendFL}{} 
\newcommand{\DIFscaledelfig}{0.5}
\RequirePackage{settobox} 
\RequirePackage{letltxmacro} 
\newsavebox{\DIFdelgraphicsbox} 
\newlength{\DIFdelgraphicswidth} 
\newlength{\DIFdelgraphicsheight} 
\LetLtxMacro{\DIFOincludegraphics}{\includegraphics} 
\newcommand{\DIFaddincludegraphics}[2][]{{\color{blue}\fbox{\DIFOincludegraphics[#1]{#2}}}} 
\newcommand{\DIFdelincludegraphics}[2][]{
\sbox{\DIFdelgraphicsbox}{\DIFOincludegraphics[#1]{#2}}
\settoboxwidth{\DIFdelgraphicswidth}{\DIFdelgraphicsbox} 
\settoboxtotalheight{\DIFdelgraphicsheight}{\DIFdelgraphicsbox} 
\scalebox{\DIFscaledelfig}{
\parbox[b]{\DIFdelgraphicswidth}{\usebox{\DIFdelgraphicsbox}\\[-\baselineskip] \rule{\DIFdelgraphicswidth}{0em}}\llap{\resizebox{\DIFdelgraphicswidth}{\DIFdelgraphicsheight}{
\setlength{\unitlength}{\DIFdelgraphicswidth}
\begin{picture}(1,1)
\thicklines\linethickness{2pt} 
{\color[rgb]{1,0,0}\put(0,0){\framebox(1,1){}}}
{\color[rgb]{1,0,0}\put(0,0){\line( 1,1){1}}}
{\color[rgb]{1,0,0}\put(0,1){\line(1,-1){1}}}
\end{picture}
}\hspace*{3pt}}} 
} 
\LetLtxMacro{\DIFOaddbegin}{\DIFaddbegin} 
\LetLtxMacro{\DIFOaddend}{\DIFaddend} 
\LetLtxMacro{\DIFOdelbegin}{\DIFdelbegin} 
\LetLtxMacro{\DIFOdelend}{\DIFdelend} 
\DeclareRobustCommand{\DIFaddbegin}{\DIFOaddbegin \let\includegraphics\DIFaddincludegraphics} 
\DeclareRobustCommand{\DIFaddend}{\DIFOaddend \let\includegraphics\DIFOincludegraphics} 
\DeclareRobustCommand{\DIFdelbegin}{\DIFOdelbegin \let\includegraphics\DIFdelincludegraphics} 
\DeclareRobustCommand{\DIFdelend}{\DIFOaddend \let\includegraphics\DIFOincludegraphics} 
\LetLtxMacro{\DIFOaddbeginFL}{\DIFaddbeginFL} 
\LetLtxMacro{\DIFOaddendFL}{\DIFaddendFL} 
\LetLtxMacro{\DIFOdelbeginFL}{\DIFdelbeginFL} 
\LetLtxMacro{\DIFOdelendFL}{\DIFdelendFL} 
\DeclareRobustCommand{\DIFaddbeginFL}{\DIFOaddbeginFL \let\includegraphics\DIFaddincludegraphics} 
\DeclareRobustCommand{\DIFaddendFL}{\DIFOaddendFL \let\includegraphics\DIFOincludegraphics} 
\DeclareRobustCommand{\DIFdelbeginFL}{\DIFOdelbeginFL \let\includegraphics\DIFdelincludegraphics} 
\DeclareRobustCommand{\DIFdelendFL}{\DIFOaddendFL \let\includegraphics\DIFOincludegraphics} 

\usepackage{tikz}
\newcommand\copyrightnotice[1]{
    \begin{tikzpicture}[remember picture,overlay]
    \node[anchor=north,xshift=-65,yshift=-8pt] at (current page.north) {\parbox{\dimexpr0.75\textwidth-\fboxsep-\fboxrule\relax}{\scriptsize #1}};
    \end{tikzpicture}
}

\usepackage[flushleft]{threeparttable}

\def\BibTeX{{\rm B\kern-.05em{\sc i\kern-.025em b}\kern-.08em
    T\kern-.1667em\lower.7ex\hbox{E}\kern-.125emX}}

\begin{document}

 \title{A Framework for Adaptive Load Redistribution in Human-Exoskeleton-Cobot Systems}

\author{Emir Mobedi$^{1}$, Gokhan Solak$^{2}$, and Arash Ajoudani$^{2}$
\vspace{-4mm}
\thanks{{$^1$Izmir Institute of Technology, Izmir, Turkiye. $^2$Istituto Italiano di Tecnologia, Genoa, Italy. This work was supported by the Horizon Europe Project Tornado under Grant Agreement No. 101189557. Email: {\tt\small{emirmobedi@iyte.edu.tr}, {gokhan.solak}@iit.it}}}%
}



\maketitle

\begin{abstract}
Wearable devices like exoskeletons are designed to reduce excessive loads on specific joints of the body. Specifically, single- or two-degrees-of-freedom (DOF) upper-body industrial exoskeletons typically focus on compensating for the strain on the elbow and shoulder joints. However, during daily activities, there is no assurance that external loads are correctly aligned with the supported joints. Optimizing work processes to ensure that external loads are primarily (to the extent that they can be compensated by the exoskeleton) directed onto the supported joints can significantly enhance the overall usability of these devices and the ergonomics of their users. Collaborative robots (cobots) can play a role in this optimization, complementing the collaborative aspects of human work. In this study, we propose an adaptive and coordinated control system for the human-cobot-exoskeleton interaction. This system adjusts the task coordinates to maximize the utilization of the supported joints. When the torque limits of the exoskeleton are exceeded, the framework continuously adapts the task frame, redistributing excessive loads to non-supported body joints to prevent overloading the supported ones. We validated our approach in an equivalent industrial painting task involving a single-DOF elbow exoskeleton, a cobot, and four subjects, each tested in four different initial arm configurations with five distinct optimisation weight matrices and two different payloads.

\end{abstract}
\vspace{-2mm}

\copyrightnotice{\copyright 2025 IEEE.  Personal use of this material is permitted.  Permission from IEEE must be obtained for all other uses, in any current or future media, including reprinting/republishing this material for advertising or promotional purposes, creating new collective works, for resale or redistribution to servers or lists, or reuse of any copyrighted component of this work in other works.}


\vspace{-5mm}
\section{Introduction}
\label{sec:sec.1}
\vspace{-2mm}

\IEEEPARstart{M}{anual} operations such as packaging \cite{napolitano20122012}, assembly \cite{03975c2e62f7423599b0da9c94c925fa} and painting \cite{rosati2014investigating} are essential in many industries, though they can place a significant strain on the physical health of human workers. Research highlights the financial strain of these issues, for instance, a study found that recovery costs for painting workers suffering from repetitive muscle strain injuries range from $\text{20,000}$ to $\text{100,000}$ USD \cite{hunting2004occupational, rosati2014investigating}. Consequently, repetitive tasks are a major contributor to work-related musculoskeletal disorders (WMSDs), which affect approximately $20\%$ of the global population \cite{glock2021assistive, ranney1995upper}.
\begin{figure}[!t]
\centering
    \includegraphics[trim=0cm 0.0cm 0cm 0cm,clip,width=.9
    \columnwidth]{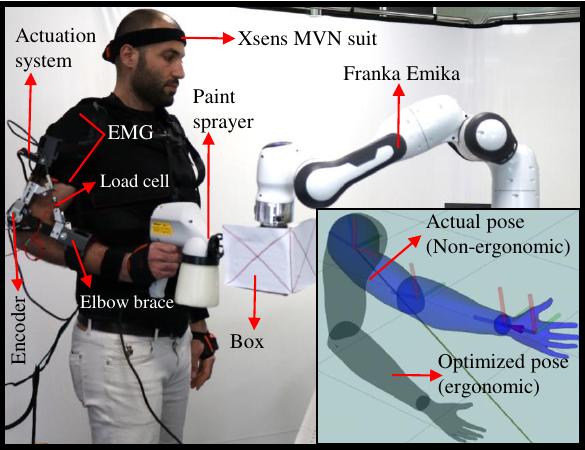}    
    \vspace{-2.7mm}
    \caption{This paper presents a solution that integrates exoskeletons for joint support, humans for supervisory roles, and cobots for adaptive task planning to optimize joint usage. It ensures that external loads are primarily directed onto the supported joints, to the extent that the exoskeleton can compensate. Through adaptive control, the cobot helps align the actual human pose (blue) with the optimized solution which is fed back to the user (grey).}
    \label{fig:Fig.1}
    \vspace{-5.0mm}
\end{figure}
\IEEEpubidadjcol

From a positive perspective, the introduction of collaborative robots (cobots) \cite{haddadin2016physical} and wearable assistive devices \cite{9435109} over the past decade has enhanced safety and productivity in operations such as drilling \cite{9982000}, assembly \cite{ZHANG2022102227}, {manual handling \cite{10684723}, helping therapists\cite{9044784}}, and pick-and-place tasks \cite{7053853}, by reducing the risk of injuries and fatigue for human co-workers \cite{peternel2019selective}. For example, in \cite{9197296}, a bi-manual human-robot cooperation task was developed for carrying a heavy box (10 kg). Optimization was applied to minimize the sum of hand contact forces with the box via the robot, and user ergonomics were evaluated using the Rapid Entire Body Assessment (REBA) method. Despite the low REBA scores indicating good ergonomics, the worker still needed to exert high torque due to the weight of the box. In \cite{8968154}, a control method was designed to reduce human fatigue when manipulating heavy objects in industrial tasks. Although this method improved user posture, repetitive tasks may still lead to WMSDs, as no additional physical support was provided to alleviate the remaining torque load. Similarly, in \cite{7987084}, a human-robot collaboration (HRC) control framework was developed for heavy material handling. The optimization minimized targeted joint torques, considering constraints such as human joint limits and robot workspace. The robot guided the human worker to an optimized body configuration; however, the method was limited to static conditions, and reliance on a force plate for joint torque estimation restricted the control framework's flexibility. Over subsequent years, this algorithm has been adapted for more dexterous tasks like drilling or polishing, incorporating additional constraints (e.g., human arm muscular manipulability and safety) to achieve an ergonomic working pose in the sagittal plane \cite{kim2021human}. Despite significant torque reduction in both studies mentioned, high effort remains in specific joints (e.g., 15 Nm in the biceps in \cite{kim2021human}). Finally, in \cite{8664488}, a control framework was developed to enable ergonomic and reconfigurable HRC. Although the online estimated torque values were optimized to guide the worker to ergonomic poses, the average optimized joint torques still posed a notable load, reported around 15 Nm in the shoulder and 12 Nm in the elbow. {Alternatively, a box lifting task is carried out with an elbow exoskeleton, transferring support up to $12$ Nm \cite{10684723}. Yet, the torque of the shoulder joint is at high level. In \cite{9044784}, a soft assistive device is developed for shoulder joints to reduce the effort of the therapists in rehabilitation tasks. In this case, no assistance is provided for elbow joint.   }
\begin{figure*}[!t]
	\centering
	\includegraphics[trim=0cm 0cm 0cm 0cm, width=.7\linewidth]{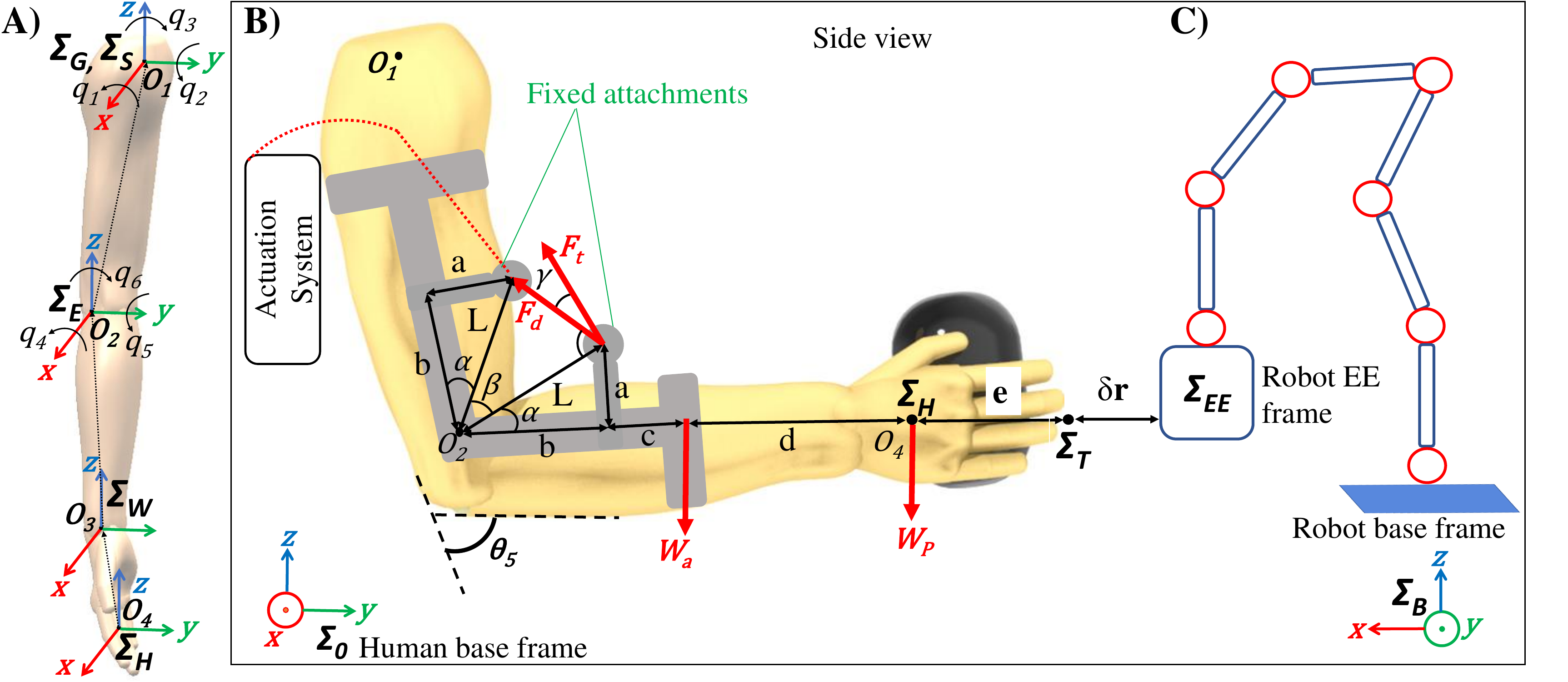}
	\vspace{-2 mm}
    \caption{A) The isometric view of the right arm with the assigned coordinates. (B) The illustration of the arm attachments from side view. The dashed red line represents the Bowden cable that transfers the generated assistive force from the actuation system to the arm. The vertical distance between elbow brace (shown in gray color) and first anchor point on the forearm (a), the distance of fixed attachments from elbow frame $\Sigma_{E}$ (b), the center of mass of the forearm (b+c), and the lever arm (b+c+d) from the external load ($W_{P}$) to $O_{2}$. (C) shows the details of the robot with the assigned coordinates from side view. (e) is the fixed distance between hand and target frame. 
    }
	\label{fig:Fig.2}
	\vspace{-5 mm}
\end{figure*}

Although the proposed solutions are promising, it is clear that neither cobots nor exoskeletons alone have achieved full minimization of the loads imposed on human body joints in collaborative tasks. Furthermore, when using exoskeletons with limited degrees of freedom (DOF), there is no assurance that external loads are properly aligned with the supported joints. For example, with an elbow exoskeleton, a task like painting may require movement in ways that place additional load on unsupported joints, such as the shoulder, which contradicts the purpose of using an exoskeleton. 

This paper proposes a solution to address these challenges by integrating the strengths of exoskeletons (for localized joint support), humans (as supervisors), and cobots (for adaptive collaborative task planning). The approach focuses on supporting individuals equipped with wearable assistive devices on specific joints, while a cobot assists in optimizing task coordination to maximize the use of the supported joints (see Fig. \ref{fig:Fig.1}). For example, if the wearable assistive device is positioned at the elbow, the cobot optimization does not need to prioritize torque minimization at that joint. Instead, it can focus more on other joints, such as the shoulder, as long as the maximum allowable torque in the supported joint remains within acceptable limits. To the best of our knowledge, this represents the first attempt to design a control framework that integrates both a robot and an exoskeleton operating together to minimize joint loading during 3D movements.

The control framework involves simplified human arm modelling, online optimization for overloading joints, ergonomic cobot planning, and the control of a 1-DOF elbow exoskeleton device \cite{10115505}. The human right arm is modeled as a $6$ DoFs rigid linkage bearing $3$ DoFs in the shoulder, and $3$ DoFs in the elbow. The human arm pose is then captured in real time using an Inertial Measurement Unit (IMU)-based system. Inverse kinematics is applied to estimate the joint angles needed for torque calculations. An online optimization technique, incorporating constraints like the robot's workspace and human joint limits, is used to distribute the interaction torques — referred to as overloading torques in this paper — arising from the weight of a handheld object between the shoulder and elbow joints. The optimized human joint angles are configurations where the overloading torques are primarily directed toward the supported elbow joint (due to the use of the elbow exoskeleton) until a threshold is reached, at which point the torques are redistributed also to the shoulder joints. Next, those joint angles are fed to the human model and displayed to the worker with the help of an avatar. Subsequently, the cobot moves the co-manipulated object in such a way to facilitate the alignment of the user's actual pose with that of the avatar (see Fig.\ref{fig:Fig.1} at the bottom right). In our work, the torque calculations are conducted under quasi-static assumptions, as manual tasks involving co-carrying or painting require slow movements due to the potential for heavy loads, as noted in \cite{7987084}.
\begin{table*}[]
\caption{INPUTS OF THE OPTIMIZATION PROBLEM}
\centering
\begin{center}  
\label{optimization parameters}
\vspace{-2mm}
\begin{tabular}{@{}cccccc@{}}
\toprule
 \multicolumn{6}{l}{\hspace{-6pt}\text{HBPs [mm]\hspace{15pt}${l_{12}}_{_{1}}, {l_{12}}_{_{2}}, {l_{12}}_{_{3}}, {l_{23}}_{_{1}}, {l_{23}}_{_{2}}, {l_{23}}_{_{3}}, {l_{34}}_{_{3}}$\hspace{15pt}$:$\hspace{15pt}$24, 57, 301, 10, 4, 296, 90$\hspace{15pt}\text{EPs [mm]\hspace{15pt}$a, b, c, d$\hspace{15pt}$:$\hspace{15pt}$50, 100, 50, 150$} 
                   }}                    \\ \midrule

 \multicolumn{6}{l}\text{\hspace{-12pt}ICs [m, rad]\hspace{10pt}$\bm{p}_{L}=[0.3 -0.8 -0.8],\hspace{6pt}\bm{p}_{U}=[0.9\hspace{5pt}0.8\hspace{5pt}0.8],$\hspace{6pt}$\bm{\theta}_{L}=[-0.22\pi -0.5\pi -0.06\pi\hspace{5pt}0-0.5\pi-0.06\pi],$\hspace{6pt}$\bm{\theta}_{U}=[0\hspace{5pt}0.5\pi\hspace{5pt}0.06\pi\hspace{5pt}0\hspace{5pt}0\hspace{5pt}0]$}                                                                                                                                                                           \\ \midrule
            & {\textbf{A}}                                                                & {\textbf{B}}                                                                & {\textbf{C}}                                                           & {\textbf{D}}                                                                        & \textbf{E}                                                                            \\\midrule
 $\bm{\theta}_{i}$ [rad] & $[0\hspace{3pt}0\hspace{3pt}0\hspace{3pt}0-0.5\pi\hspace{3pt}0]$ & $[0\hspace{3pt}0\hspace{3pt}0\hspace{3pt}0-0.5\pi\hspace{3pt}0]$ & $[0-0.4\pi\hspace{3pt}0\hspace{3pt}0-0.15\pi\hspace{3pt}0]$ & $[0-0.35\pi\hspace{3pt}0\hspace{3pt}0-0.15\pi\hspace{3pt}0]$ & $[-0.3\pi\hspace{3pt}-0.2\pi\hspace{3pt}0\hspace{3pt}0-0.5\pi\hspace{3pt}0]$ \\
$\mathbb{W}$ & $(0, \hspace{3pt}0, ..., 0)$                                     & $(0.2,\hspace{3pt}1, ..., 0)$                                    & $(0,\hspace{3pt}1, ..., 0)$                                 & $(0.2,\hspace{3pt}1 ..., 0)$                                             & $(0.2,\hspace{3pt}1 ..., 0)$                                                 

\\ \bottomrule
\end{tabular}
\end{center}
\vspace{-2mm}
\begin{tablenotes}
\item{Human Body Parameters: HBPs,  Exoskeleton Parameters: EP, Inequality Constraints: ICs,
Cartesian position vector: $\bm{p}$, Joint angles vector: $\bm{\theta}$.}
\vspace{-6mm}
\end{tablenotes}
\end{table*}

As shown in Fig.\ref{fig:Fig.1}, the actuation mechanism of the assistive device is attached to the back of the human worker, and its support force is transmitted to the elbow brace through the Bowden cable. The essential goal of the assistive mechanism is to alleviate the muscular effort of the human because of internal (i.e., $2$ kg forearm weight in \cite{cappello2015series}, \cite{dempster1967properties}), and external (lightweight tools up to $1$ kg) loading. This approach prevents prolonged exposure to repetitive loading, leading to fatigue, thereby enhancing the productivity of robotic co-workers in industrial settings. Additionally, force control is incorporated to leverage the human's supervisory capabilities. Specifically, the controller takes the elbow position as an input and outputs the assistive torque applied to the elbow joint. This enables the user to move his/her arm freely without encountering opposing forces, guiding it to an optimized pose with minimal physical or control-based constraints during the operation.

To show the applicability and the performance of the control technique, the verification is performed for a painting task on $4$ subjects under $4$ different arm configurations with different weight matrix defined in the optimization, and different payloads. Furthermore, the tests are conducted with and without exoskeleton assistance while measuring the effort through an electromyography interface in the shoulder and elbow.

The rest of the paper is structured as follows. In Section \ref{sec:sec.2}, the methodology comprising human-arm modeling and its static force analysis, online optimization, robot planning, and exoskeleton control are studied. In Section \ref{sec:sec.3} the experiments and the results are presented. Finally, in Section \ref{sec:sec.4}, the conclusions and future work are highlighted.


\vspace{-4.0mm}
\section{Methodology}
\label{sec:sec.2}
\vspace{-1mm}
The goal of this section is to introduce the details of the control framework including the static force analysis of the human right arm, online optimization for the minimization of the overloading joint torques, robot path planning, and the exoskeleton assistance. 

\vspace{-5.5mm}
\subsection{Human Arm Model}
\label{ssec:Human model}
\vspace{-2mm}
The human right arm is modeled in URDF (Unified Robotics Description Format) as a $6$ DoFs rigid link $3$ DoFs of which are assigned for the shoulder ($\Sigma_{S}$), and the elbow ($\Sigma_{E}$) as in Fig.\ref{fig:Fig.2}A. Although we have $2$ DoFs anatomically for the elbow including flexion/extension and supination/pronation, the XSENS human monitoring system \cite{CiteDrive2022} assigns $3$ DoFs for each joint based on ISB biomechanical model \cite{WU2005981}. Thus, it is mimicked in our URDF model.

To begin with, $\Sigma_{G}$ is the global frame, and located exactly at the same point as $\Sigma_{S}$ to compute the torques/forces of other frames with respect to $\Sigma_{G}$. On the other hand, $\Sigma_{W}$, and $\Sigma_{H}$ that are the wrist and hand frame are not considered as a joint. Instead, those two frames are defined as a fixed point for the forward kinematic calculations of hand with respect to $\Sigma_{G}$. The position vectors between $\Sigma_{S}$, $\Sigma_{W}$, $\Sigma_{E}$, and $\Sigma_{H}$ are stated as follows:
\vspace{-3mm}
\begin{equation}
\bm{{O_{1}O_{2}}}=({l_{12}}_{_{1}})\bm{x}-({l_{12}}_{_{2}})\bm{y}-({l_{12}}_{_{3}})\bm{z},
\label{eq:1}
\vspace{-2mm}
\end{equation}
\vspace{-2mm}
\begin{equation}
\bm{{O_{2}O_{3}}}=-({l_{23}}_{_{1}})\bm{x}-({l_{23}}_{_{2}})\bm{y}-({l_{23}}_{_{3}})\bm{z},
\label{eq:2}
\vspace{-2mm}
\end{equation}
\vspace{-2mm}
\begin{equation}
\bm{O_{34}}=-({l_{34}}_{_{3}}) \bm{z},
\label{eq:3}
\vspace{-2mm}
\end{equation}
where ${l_{ij}}_{_{h}}$ denotes the distance between $i^{th}$ and $j^{th}$ points along $h^{th}$ principal axis. For instance, $\bm{{O_{1}O_{2}}}$, $\bm{{O_{2}O_{3}}}$, $\bm{{O_{3}O_{4}}}$ are the position vectors, and their numerical values are illustrated in Table.\ref{optimization parameters}. First one represents the location of elbow joint relative to shoulder. Second vector shows the location of elbow joint with respect to wrist frame, and the last one defines the position of wrist with respect to hand frame. 

In the next step, the dynamic model of the arm in joint space can be written based on Lagrangian formulation. The generalized coordinates are defined as $\bm{\theta}\in \mathbb{R}^n$, for $n$-DoFs arm. Thus:
\vspace{-5mm}
\begin{equation}
    \bm{B}(\bm{\theta})\ddot{\bm{\theta}}+\bm{C}(\bm{\theta},\dot{\bm{\theta}})\dot{\bm{\theta}} +\bm{g}(\bm{\theta}) = \bm{\tau} - \overbrace{{\bm{J}(\bm{\theta})^T\bm{F}_{ext}}}^{\bm{\tau}_{ext}},
\label{eq:4}
\vspace{-2mm}
\end{equation}
where $ \bm{B}(\bm{\theta})$, and $\bm{C}(\bm{\theta},\dot{\bm{\theta}})$ are $n\times n$ inertia and coriolis/centrifugal matrix, respectively. In addition, $\bm{g}(\bm{\theta})$ is the vector of gravity joints torques, and $\bm{\tau}$ is the vector of joint torques human applies to achieve the motion. $\bm{F}_{ext}\in \mathbb{R}^{k}$ is the external forces/moments vector, and its effect in joint space is computed via $\bm{\tau}_{ext}$. Finally, the Jacobian matrix is defined as $\bm{J}(\bm{\theta}) \in \mathbb{R}^{k\times n}$, where $k=6$ at the human hand in Cartesian space, and $n=6$ in joint space.

As previously mentioned, since the control framework is designed for quasi-static movements (i.e., slow motions), a simplified approach for calculating overloading joint torques (i.e., torque values imposed on human arm joints due to external interactions) is applied as follows
\vspace{-2mm}
\begin{equation}
\bm{\tau}=\bm{J}(\bm{\theta})^T\bm{F}_{ext},
\label{eq:5}
\vspace{-1mm}
\end{equation}
where the components of $\bm{F}_{ext}$ can be measured through the cobot’s Cartesian forces, with an additional force in the $z$ direction accounting for any object weight held by the user. Here, each component of the forward kinematics function for translational, and rotational movement is specified by the $x,y,z$, and $\varphi, \vartheta, \psi$ (RPY), respectively. As presented in Fig.\ref{fig:Fig.2}B, the only external force occurs along $-z$ axis due to the tool ($W_{P}$) user holds. The forearm weight (i.e., internal loading) represented with $W_{a}$ is not taken into account here since it is fully compensated with the help of an exoskeleton support that will be mentioned later.
\vspace{-5mm}
\subsection{Online Optimization}
\label{ssec:Optimization} 
\vspace{-1.2mm}
In this chapter, online optimization formulation is studied to obtain an arm configuration such that the overloading joint torques are minimized. In the optimization problem, motion range of the joints, and cartesian movement limits are determined as the inequality constraints. Next, the cost function is built, and Frobenius norm $|| . ||$ is utilized to find its local minima. Therefore: 
\vspace{-4mm}

\begin{mini}
{\theta}{f_{0}(\bm{\theta})=||\bm{\tau}^T\mathbb{W}\bm{\tau}||}
{\label{eq:6}}{}
\end{mini}
\vspace{-3mm}
$$
\text{subject to}\left\{
    \begin{array}{ll}
    {f_{1}(\bm{\theta})=\bm{\theta}_{L}-\bm{\theta}\leq 0,\hspace{8pt} f_{3}(\bm{\theta})=\bm{p}_{L}-\bm{r}\leq 0}\\
    {f_{2}(\bm{\theta})=\bm{\theta}-\bm{\theta}_{U}\leq 0,\hspace{8pt} f_{4}(\bm{\theta})=\bm{p}-\bm{p}_{U}\leq 0}\\
\end{array}
\right.
\vspace{-1mm}
$$
where $\bm{p}\in \mathbb{R}^{k\times 1}$ is the vector of positions in task space, and defined within the robot workspace. $\mathbb{W}\in \mathbb{R}^{k\times n}$ is a symmetric positive definite weight matrix, and each diagonal member of it \eqref{eq:7} corresponds to the joint numbers. Moreover, $\bm{\theta}_{L},\bm{p}_{L}$, and $\bm{\theta}_{U},\bm{p}_{U}$ signify the lower and upper boundary of the inequality constraints, respectively.
\vspace{-2mm}
\begin{equation}
\mathbb{W}=\text{diag} (W_{1}, W_{2},..., W_{n}).
\label{eq:7}
\vspace{-2mm}
\end{equation}
Augmented Lagrangian solver (AUL) of ALGLIB library is employed for the minimization of the objective function. In this method, first gradient of the functions including objective and inequality constraints should have non-zero values at the determined boundary.
By assigning different weights to the shoulder and elbow joints, various arm configurations can be achieved. The choice of weights depends on which joints require minimized overloading torques. For instance, in the case of the elbow exoskeleton, the weight will remain zero until the exoskeleton reaches its full torque capacity. The range of weight is also important during the tuning process. It is stated that the relative differences among weights are more important than their independent absolute values \cite{marler2010weighted}. Hence, the range is specified between $0$ and $1$. For instance, specifying $W_{2} \approx 1$ and $W_{5} \approx 0$ generates a similar arm configuration as in Fig. \ref{fig:Fig.2}B, that is, minimum torque at the shoulder ($\theta_{2} \approx 0^{\circ}$), and max torque at the elbow ($\theta_{5} \approx -90^{\circ}$). More discussion and concrete results about the effects of weight on the cost function are presented in \ref{sec:sec.3}.
\vspace{-0.54cm}
\subsection{Robot Planning}
\label{ssec:Robot Planning}
\vspace{-0.1cm}
Once an optimal configuration for the human arm is achieved, the robot's path planner aims to help the user reach a similar arm configuration. Simultaneously, real-time visual feedback is provided to the user via an avatar displaying both the optimized pose and the actual pose, as shown in Fig.\ref{fig:Fig.1}. The user then adjusts their arm to align the actual pose with the optimized one, with the robot following this movement. During this process, the robot tracks the human hand, allowing the painting task to be completed with reduced effort and under the user's supervision. To enable this, a target frame $\Sigma_{T}$ is defined with a translational offset ($e$) relative to the hand frame (see Fig.\ref{fig:Fig.2}). The robot is then controlled to follow this target along a linear trajectory, calculated based on the reaching time and the positional error between the target and the robot's end-effector frame ($\delta\bm{r}$). A Cartesian impedance controller is developed for the trajectory tracking, and to achieve a compliant behaviour at the robot end-effector.  
\begin{figure*}[!t]
\centering
    \includegraphics[trim=0cm 0.0cm 0cm 0cm,clip,width=1.5
    \columnwidth]{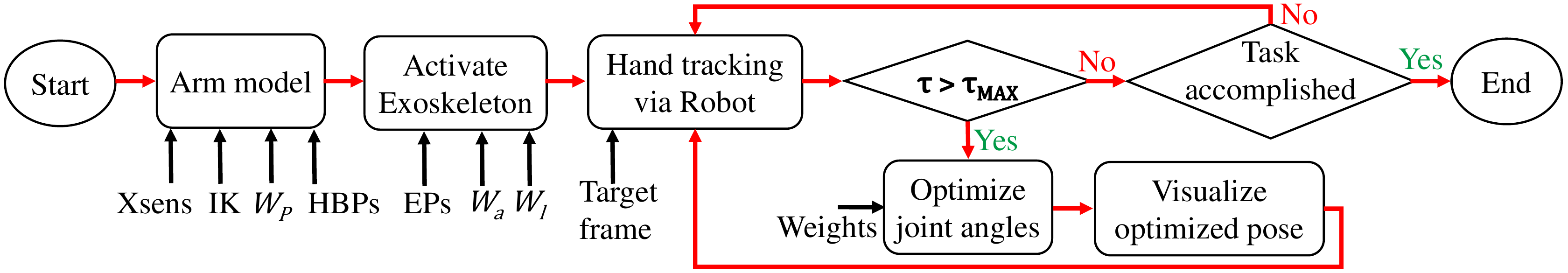}    
    \vspace{-2.2mm}
    \caption{The proposed control strategy to alleviate the overloading joint torques by means of the robot and the elbow exoskeleton. IK and HBPs represent the inverse kinematics, and human body parameters, respectively. Black arrows are the inputs whereas red arrows show the flow of the control algorithm.    }
    \label{fig:Fig.3}
    \vspace{-5.0mm}
\end{figure*}
\vspace{-1cm}
\subsection{Exoskeleton Assistance}
\label{ssec:Exoskeleton support}
\vspace{0cm}
Even though optimization of the joint angles can reduce the overloading joint torques through the assistance of the robot, the remaining effort on the user's joints is still not trivial due to the external and internal loading's. To amplify the physical support to the worker, an elbow assistive device is attached to the user without preventing the arm movements in $3$D space during the robot path planning. The actuation system of the device is developed through series elastic actuation (SEA) principle. An endless shape bungee element is elongated linearly by the ball-screw and a DC motor (ECXTQ22XL, GPX22UP) to generate the desired force in a compliant way. The support that device provides cover the entire forearm weight ($W_{a}$) and some portion of the external load ($W_{P}$) as shown in Fig.\ref{fig:Fig.2}.
The assistive torque ($\tau_{Exo}$) about $O_{2}$ is computed as follows:  
\vspace{-1.5mm}
\begin{equation}
\tau_{Exo}=\overbrace{{W}_{l}{\sin({\theta_{5}})}{l_{L}}}^{\tau_{L}}+\overbrace{{{W}_{a}}{\sin({\theta_{5}})}l_{A}}^{\tau_{A}},\hspace{0.3cm}{{F}_R}=\cfrac{\tau_{Exo}}{L\cos{(\gamma)}} 
\label{eq:10}
\vspace{-1mm}
\end{equation}
where $l_{A}=b+c$, and $l_{l}=b+c+d$ are the lever length for the forearm and the external load, respectively. $W_{l}$ is the external load that the device can fully compensate, and it is smaller than the actual load ($W_{P}$). The force trajectory to transfer this torque to the elbow joint is represented with $F_{R}$, which includes some geometrical relations of the arm attachments, forearm weight, external load, and joint angle as the variable. Thus, the motion change on the elbow joint is captured via an encoder and $F_{R}$ is updated while measuring the tendon force by means of a load cell (FUTEK-LSB201). The resultant error is fed to a PID controller as input, and the output is computed as the new bungee elongation. More details about the design and control of the mechanism can be found in \cite{10115505,10143312}.


\vspace{-0.40cm}
\section{Experiments \& Results}
\label{sec:sec.3}
An equivalent industrial painting task (see Fig.\ref{fig:Fig.1}) is designed to evaluate the developed control framework on $4$ young healthy subjects ($30\pm5$ years old), and the experiments were conducted at the Human-Robot Interfaces and Interaction (HRII) Lab, Istituto Italiano di Tecnologia (IIT), and approved by the ethics committee Azienda Sanitaria Locale (ASL) Genovese N.3 (Protocol IIT HRII SOPHIA 554/2020). 

The flowchart of the experiments is presented in Fig.\ref{fig:Fig.3}. First, the XSens MVN suit is worn by the subjects for full-body motion capture as in Fig.\ref{fig:Fig.1}. However, since our interest is to monitor only upper-body motions, $11$ miniature inertial measurement units (IMU) are used in the experiments. The frame locations of human joints w.r.t $\Sigma_{0}$ are acquired in ROS (Robot Operating System) through the software framework \cite{leonori2023bridgexsensmotioncapturerobot}. Afterward, a closed loop inverse kinematic (CLIK) algorithm \cite{8700380} is implemented to estimate the joint angles in their local frame. Next, those joint angles and human body parameters (HBPs in Table.\ref{optimization parameters}) are fed to our URDF model, where the global frame is attached in the shoulder (see Fig.\ref{fig:Fig.2}A), to estimate the joint torques through~\eqref{eq:5}. After that, those torques are employed in the cost function together with the weight matrix to minimize the overloading joint torques. As the exoskeleton compensates the torque load for elbow flexion/extension movement ($\theta_{5}$), $W_{5}$ element of the weight matrix is set always $0$. In addition,  minimizing $\theta_{3}, \theta_{4}$, and $\theta_{6}$ is not our objective, as it is not relevant to our specific use cases within the context of a European project. Additionally, \cite{rosati2014investigating} demonstrates that the primary loaded joints in a painting task are $\theta_{1}$, $\theta_{2}$, and $\theta_{5}$. Thus $W_{3}, W_{4}$, and $W_{6}$ are assigned as $0$ as in Table.\ref{optimization parameters}. However, $W_{1}$, and $W_{2}$ are varied in each test. As mentioned in \ref{ssec:Optimization}, to achieve a different priority in the optimization, it is important to evaluate the ratio between the elements of $\mathbb{W}$ rather than their absolute values. Therefore, $W_{1}$, and $W_{2}$ are determined as $0.2$, and $1$ (or vice versa) among the tests
to monitor how $\theta_{1}$, and $\theta_{2}$ are changed as a result of the optimization process. Moreover, the inequality constraints are defined for the position limits in task space ($\bm{r_{L}, r_{U}}$) and the joint angle limits ($\bm{\theta_{L}, \theta_{U}}$) based on the painting task requirements. Also, the optimization is performed at different initial arm configurations represented in $\bm{\theta_{i}}$.
 \begin{figure}[!t]
\centering
\includegraphics[ trim=0cm 0cm -1cm 0cm, width=.88
\linewidth]{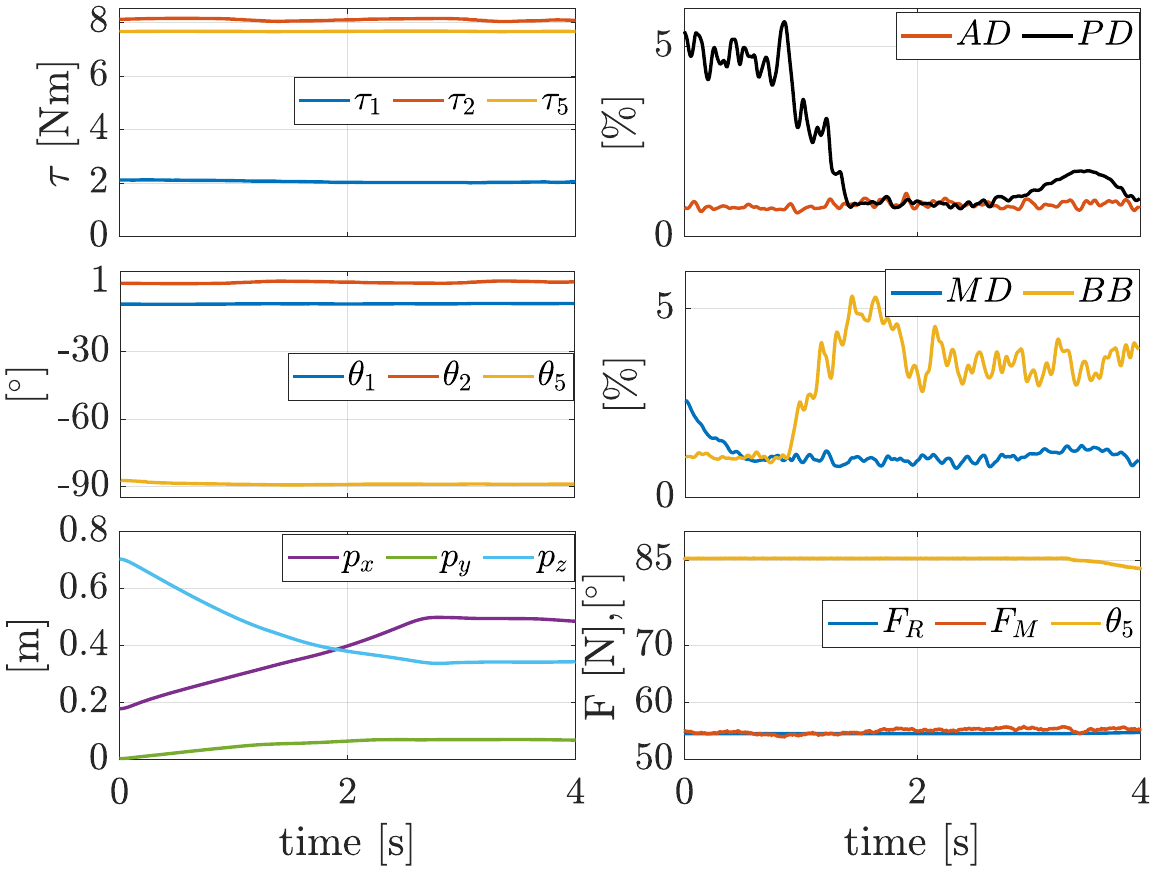}
  \vspace{-2mm}
    \caption{The experiment-A results of $S_{1}$ where $2$ kg payload is held by assigning null value in $\mathbb{W}$ to record $\bm{\tau_{Max}}$ at a predetermined pose. $F_{R}$, and $F_{M}$ are the desired and measured force of the exoskeleton, respectively. }
    \label{fig:Fig.4}
  \vspace{-5.35mm}
\end{figure}   
\begin{figure}[!b]
 \vspace{-7.5mm}
\centering
\includegraphics[ trim=0cm 0cm -10.8cm 0cm, width=1.45
\linewidth]{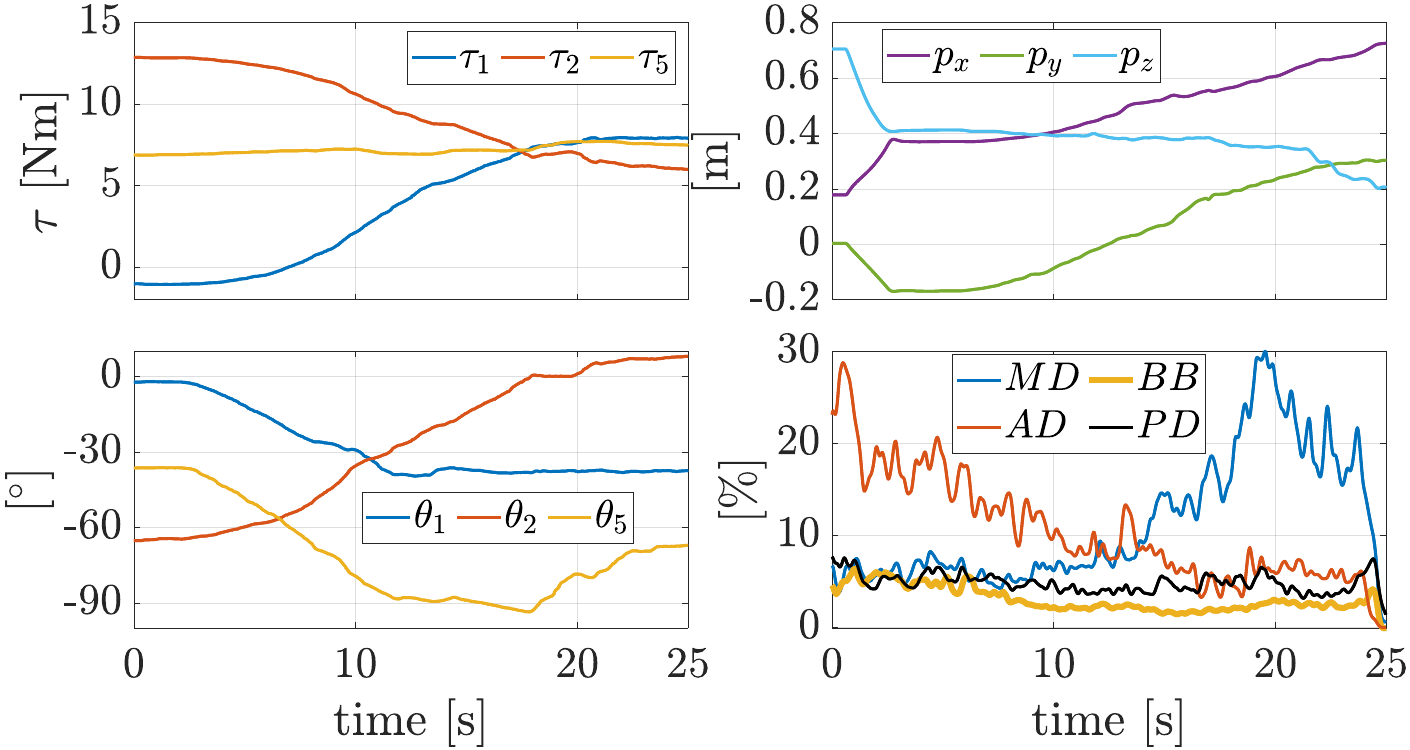}
  \vspace{-5mm}
 \caption{The experiment-C results of $S_{3}$ in which $W_{P}=2$ kg payload is held without wearing the assistive device. $F_{R}$, and $F_{M}$ are the desired and measured force of the exoskeleton, respectively. $\tau$ is the joint torque.}
  \label{fig:Fig.5}
\end{figure}
Next, the Franka Emika Panda robot is operated through a custom ROS controller, and a calibration session is performed. To map the hand movements (defined in $\Sigma_{0}$) in $\Sigma_{B}$, $\Sigma_{EE}$ and $\Sigma_{H}$ are closely coincided at the calibration pose in sagittal plane which corresponds to keeping their fist below the $\Sigma_{EE}$ around $\theta_{2} \approx 0^{\circ}$ and $\theta_{5} \approx -90^{\circ}$ (all the other joints are $\approx 0^{\circ}$). The measured error between $\Sigma_{EE}$ and $\Sigma_{H}$ is adjusted to be almost zero by changing the $\Sigma_{B}$ in RVIZ (ROS graphical interface), and participants are asked to stay at the same location throughout the experiment. After this calibration, the target frame ($\Sigma_{T}$) is created $e=20$ mm ahead of the $\Sigma_{H}$, and tracked by the robot in translational movements. The assigned stiffness gains for the cartesian impedance controller are ${K}_x={K}_{y}={K}_{z}=600\,\mathrm{N/m},  {K}_{roll}={K}_{pitch}={K}_{yaw}=30\,\mathrm{Nm/rad}$, and damping gains are set proportionally. 

Regarding the assistive device, first, the parameters including $a$, $b$, $c$, and $d$ (see Table.\ref{optimization parameters}) are set for all subjects, and a PID based force control ($K_{P_{f}}=0.1, K_{I_{f}}=6, K_{D_{f}}=0.002$) is implemented in MATLAB\textsuperscript \textregistered /Simulink Real-Time interface to track the force reference generated based on \eqref{eq:10}. Also, a data acquisition card and a motor driver are communicated via an EtherCAT at 1 kHz. In the experiments, Wagner\textsuperscript \textregistered 56652/2017 paint sprayers ($W_{P}=2$ and $4$ kg) are held by the subjects as the external load. In \eqref{eq:10}, the forearm weight is set as $20$ N for all subjects considering \cite{cappello2015series} whereas the external load ($W_{l}$) is assigned to be $10$ N (max permissible external load value). 
 \begin{figure}[!t]
\centering
\includegraphics[ trim=0cm 0cm -1cm 0cm, width=.95
\linewidth]{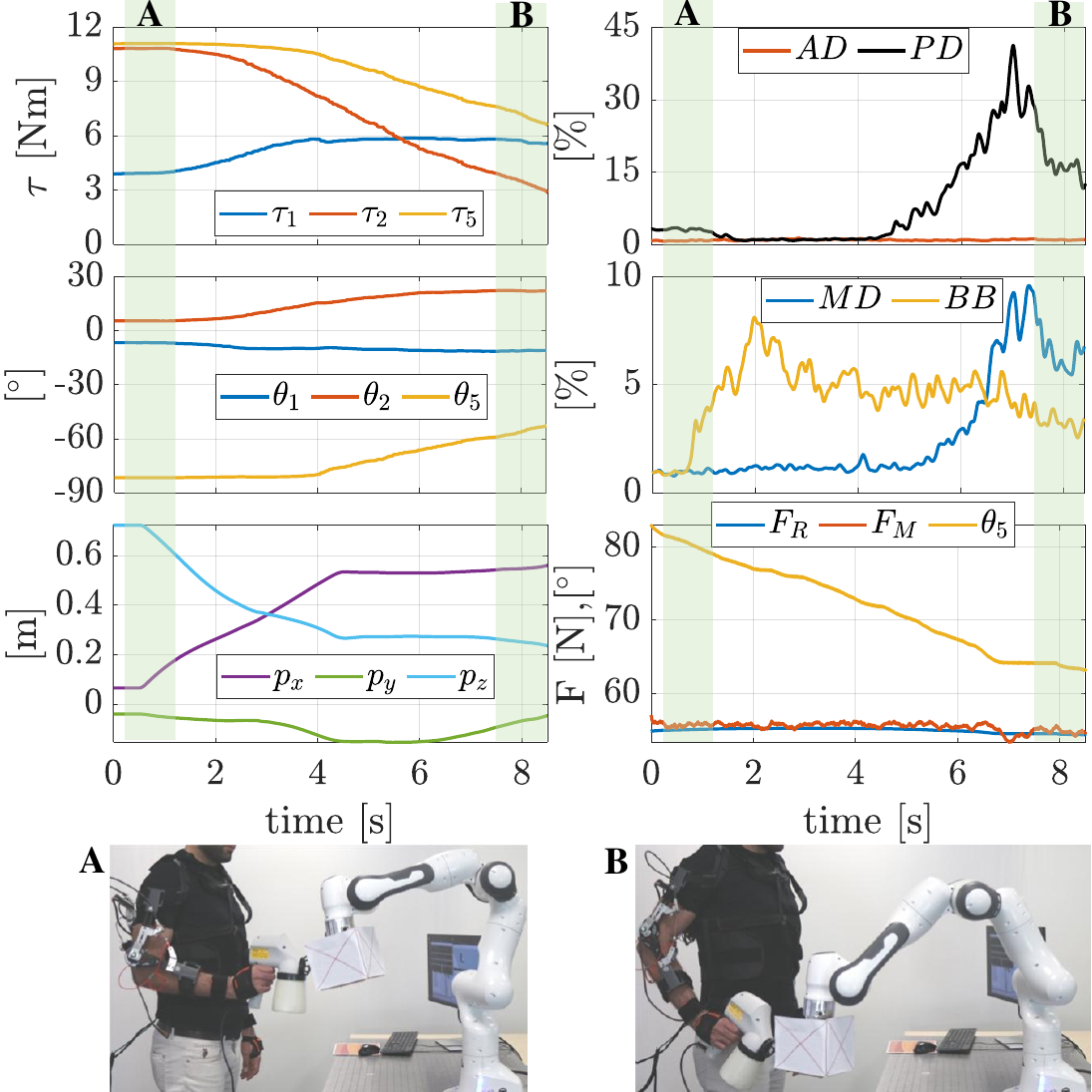}
  \vspace{-2.0mm}
    \caption{The experiment-B results of $S_{1}$ where $4$ kg payload is held by assigning nonzero $W_1$, and $W_{2}$. $F_{R}$, and $F_{M}$ are the desired and measured force of the exoskeleton, respectively. $\tau$ is the joint torque.}
    \label{fig:Fig.6}
  \vspace{-4.6mm}
\end{figure}
\begin{figure}[!b]
 \vspace{-4mm}
\centering
\includegraphics[ trim=0cm 0cm -10cm 0cm, width=1.3
\linewidth]{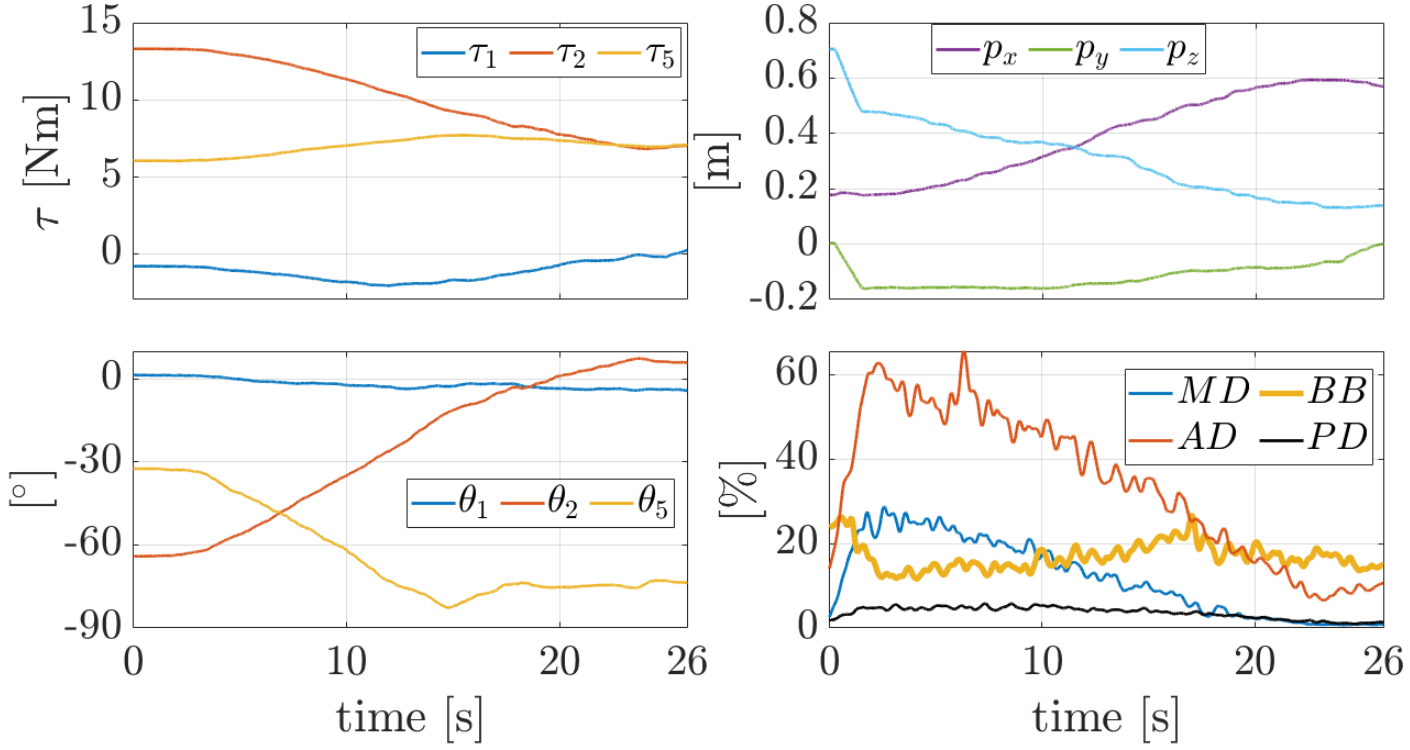}
  \vspace{-5mm}
    \caption{The experiment-D results of $S_{2}$ in which $W_{P}=2$ kg payload is held without wearing the assistive device. $F_{R}$, and $F_{M}$ are the desired and measured force of the exoskeleton, respectively. $\tau$ is the joint torque. }
    \label{fig:Fig.7}
  \vspace{0mm}
\end{figure}
During the experiments, to monitor the effort reduction on shoulder (flexion/extension and abduction/adduction) and elbow joint movements (flexion/extension), the muscular activities of anterior deltoid (AD), middle deltoid (MD), posterior deltoid (PD), and biceps brachii (BB) are recorded via the wireless sEMG device of Delsys Trigno platform by Delsys Inc (Natick, MA, United States). The EMG raw data is obtained at 2 kHz, then it is post-processed in MATLAB\textsuperscript \textregistered applying a second order high-pass filter with 0.1 Hz. Finally, full-wave rectification is carried out, and 2.5 Hz second-order low pass filter is employed on the data. 
\vspace{-6.0mm}
\subsection{Experimental Protocol}
 \label{ssec:Exp protocol}
\vspace{-1.0mm}

Each subject participated in $8$ tests, with the optimization parameters presented in Table \ref{optimization parameters}. Experiments A and B were conducted with the exoskeleton, while C, D, and E were carried out both with (W/) and without (W/O) the exoskeleton. The test order was randomized as in \cite{schalk2022influence}. Additionally, while a weight of $W_{P}=4$ kg was used in experiment B, a weight of $W_{P}=2$ kg served as the external load for all other tests.

In each experiment, subjects first stayed at the calibration location with the cobot. They were then instructed to position their arm to the initial pose (within a tolerance of $\pm0.03\pi$) shown as $\bm{\theta}_{i}$ in Table 1. Afterward, the cobot was activated and moved to $\Sigma_{T}$ to track the user’s hand until the actual arm pose aligned with the optimized pose (avatar). For tests conducted with assistance, the steps described above were completed with exoskeleton support. Additionally, a tolerance was introduced for the error between the optimized angle of the target joint (i.e., $\theta_{1}$, $\theta_{2}$, $\theta_{5}$) and the actual joint angle, set to $\theta_{E}=10^\circ\pm5^\circ$, to account for possible inconsistencies in movement coordination among subjects.
\begin{figure*}[!t]
\centering
    \includegraphics[trim=0cm 0.0cm 0cm 0cm,clip,width=1.8
    \columnwidth]{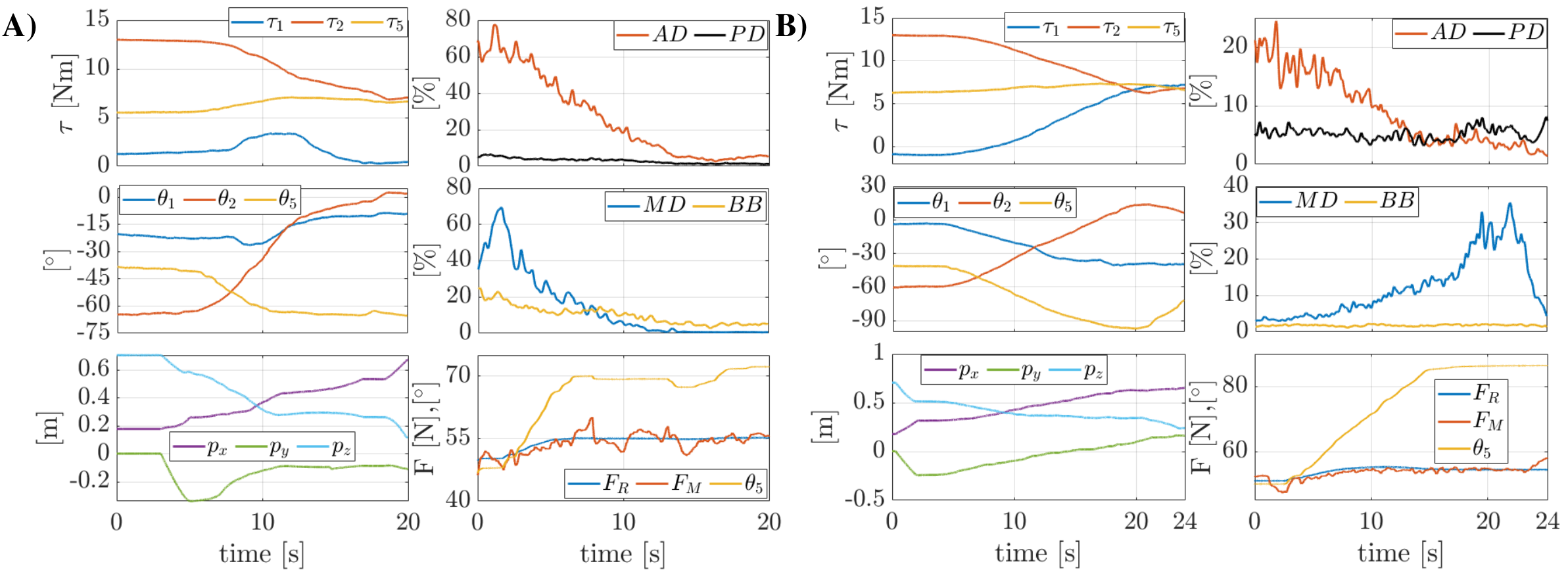}    
    \vspace{-3mm}
    \caption{A) The experiment-D results of $S_{2}$, and B) the experiment-C results of $S_{3}$. Both tests are conducted under exoskeleton support by holding $W_{P}=2$ kg. $F_{R}$, and $F_{M}$ are the desired and measured force of the exoskeleton, respectively. $\tau$ is the joint torque.}
    \label{fig:Fig.8}
    \vspace{-5mm}
\end{figure*}
In experiment-A, the $\mathbb{W}$ is defined as null, and thus no solution is expected from the optimization algorithm. The goal is to acquire the maximum allowable joint torques ($\bm{\tau_{Max}}$) whose surpassing leads to trigger the optimization for the following tests. In experiment-B, the external load is doubled at the same $\bm{\theta_{i}}$ as the experiment-A, and hence the $\bm{\tau_{Max}}$ is exceeded. Here, the optimization is carried out by determining $W_{2}$ greater than $W_{1}$ since the aim is to distribute the torque load among $\theta_{2}$, and $\theta_{5}$ while alleviating $\tau_{1}$. In experiment-C, $\bm{\theta}_{i}$, and $W_{P}$ are changed while assigning only a nonzero $W_{2}$ to monitor if the optimization can reduce the effort about only $\theta_{2}$. When it comes to experiment-D, in addition to $W_{2}$, $W_{1}$ is assigned to observe the torque reduction about both $\theta_{1}$, and $\theta_{2}$ with almost similar $\bm{\theta}_{i}$ as previous test. Finally, experiment-E is carried out by making use of the same weight matrix as experiment-D, yet the $\bm{\theta}_{i}$ is modified to amplify the initial torque of the shoulder about $\theta_{1}$, and $\theta_{2}$.
\vspace{-5.0mm}
\subsection{Experimental Results}
\label{ssec:experimental results}
\vspace{-1.0mm}

The average value of all subjects' data with standard deviation (std) for experiments B, C, D, and E are reported in Table.II. Since experiment-A is carried out only to determine the $\bm{\tau_{Max}}$ values without assigning a nonzero weight in the optimization, its results are excluded here. To start with, $\theta_{1E}$, $\theta_{2E}$, and $\theta_{5E}$ represent the error between the optimized and the actual pose of the corresponding joints, and they are reported to monitor how close the subjects' arm pose to that of the provided optimized one. They are calculated by taking the average value of the last $3$ seconds of the corresponding joint angle (i.e., after optimization) for each test, and subtracted from that of the optimized angle.
\begin{figure}[!b] 
\vspace{-7mm}
\includegraphics[ trim=0cm 0cm -1cm 0cm, width=1
\linewidth]{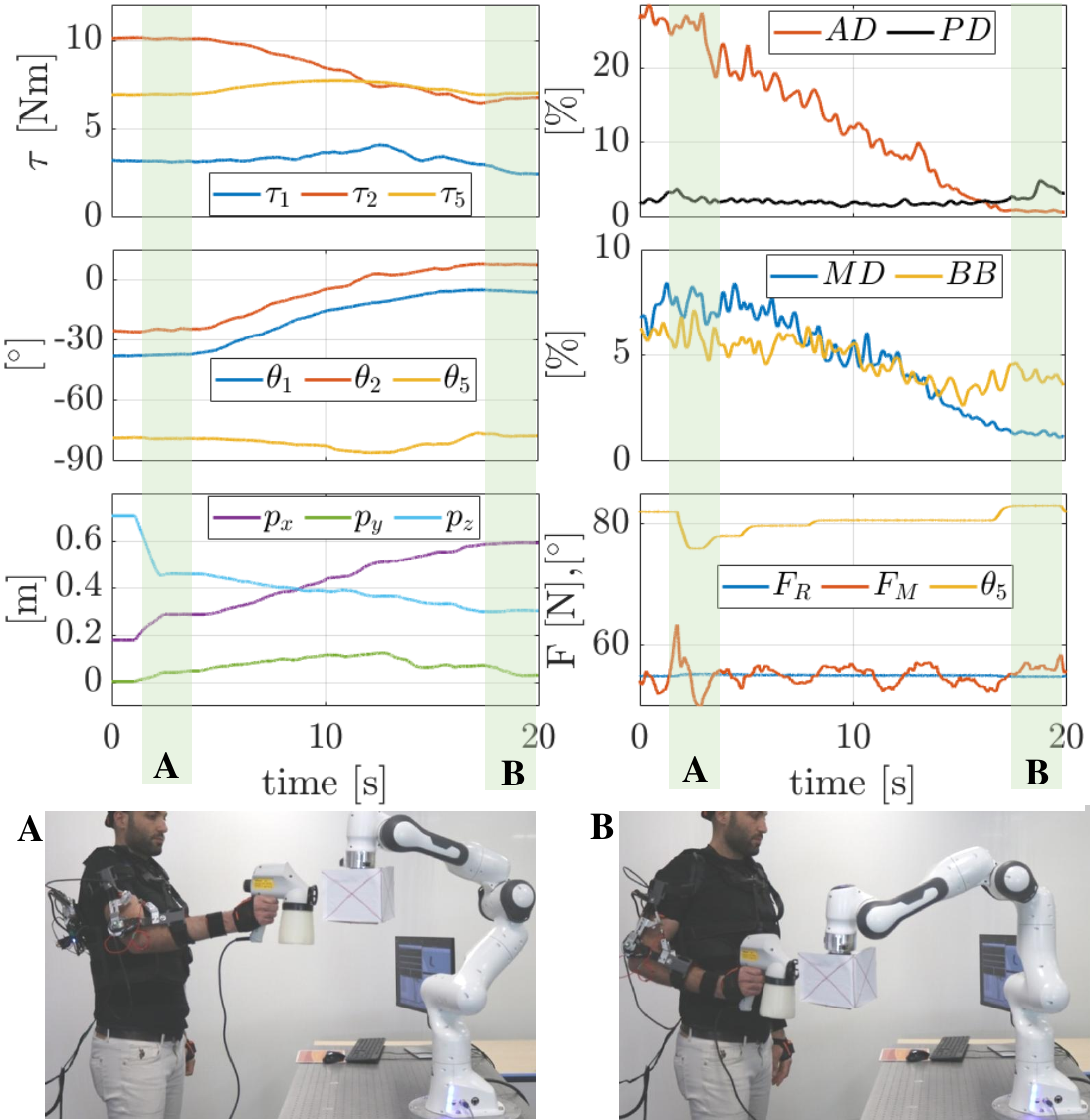}
 \vspace{-7mm}
    \caption{The experiment-E results of $S_{1}$ in which $2$ kg payload is held by assigning nonzero $W_1$, and $W_{2}$. $F_{R}$, and $F_{M}$ are the desired and measured force of the exoskeleton, respectively. $\tau$ is the joint torque.}
    \label{fig:Fig.9}
 \vspace{-1mm}
\end{figure}
Next, $F_{E}$, and $P_{E}$ are the RMS value of the force error of the exoskeleton, and the Cartesian error in $x, y, z$ axes of the robot, respectively. The latter is computed taking the RMS value of the error along each axis during the test, and then the RMS value of the error for all axes is computed. Moreover, the rate of change of torques in percentage for $\theta_{1}$, $\theta_{2}$, $\theta_{5}$ are reported by taking the average of the first and last $3$ seconds of the data as illustrated in Fig.\ref{fig:Fig.9} with state-A and B. Then, the torque difference between the two states is divided to the initial torque, and multiplied by $100$ to compute the reduction or increment percentage with respect to the initial state. However, there is an exception that is employed for $4$ tests pointed with $\dagger$ on the Table.II. To clarify, since the final $\tau_{1}$ value of those tests are drastically higher than that of the initial one (i.e., the initial condition of $\theta_{1}$ is $0$ for these tests), their rate of change is computed considering the final state to make the data interpretation easier. Furthermore, to monitor the effort change among the tests, RMS values of EMG channels including AD, MD, PD, and BB are computed for each test.

To begin with, according to the results of experiment-A in Fig.\ref{fig:Fig.4}, the $\bm{\tau_{Max}}$ are recorded as $\tau_{1}=2.02$ Nm, $\tau_{2}=8.13$ Nm, $\tau_{5}=7.66$ Nm. No movement is performed by $S_{1}$ since a null $\mathbb{W}$ is set. Considering the exoskeleton support ($F_{R}\approx 55$ N generates $\approx6$ Nm assistive torque at elbow joint based on \eqref{eq:10}), it is clear that the resultant $\tau_{5}$ is quite low.
%
\begin{table*}[]
\centering
\small
\begin{center}
\caption{{ EXPERIMENT RESULTS.} { \bf The value is reported as: mean (standard deviation)} }
\vspace{-2.0mm}
\begin{tabular}{@{}lcccccccc@{}}
\toprule
\multicolumn{1}{c}{} & \multicolumn{2}{c}{\textbf{B}} & \multicolumn{2}{c}{\textbf{C}} & \multicolumn{2}{c}{\textbf{D}} & \multicolumn{2}{c}{\textbf{E}}  \\ \midrule
\multicolumn{1}{c}{} & \textbf{W/}               & \textbf{W/O} & \textbf{W/}              & \textbf{W/O}              & \textbf{W/}              & \textbf{W/O}              & \textbf{W/}               & \textbf{W/O}              \\
$\theta_{1E}$        & $8.6\hspace{2pt}(4.2)$    & -            & $2.6\hspace{2pt}(2.5)$   & $1.7\hspace{2pt}(0.8)$    & $6.5\hspace{2pt}(2)$     & $4.3\hspace{2pt}(2.6)$    & $7\hspace{2pt}(2.4)$      & $5.8\hspace{2pt}(6.1)$    \\
$\theta_{2E}$        & $4.8\hspace{2pt}(2.7)$    & -            & $6.1\hspace{2pt}(6.2)$   & $8.9\hspace{2pt}(4.1)$    & $6.2\hspace{2pt}(3.9)$   & $9.8\hspace{2pt}(5.1)$    & $11.9\hspace{2pt}(1.1)$   & $9.1\hspace{2pt}(5.3)$    \\
$\theta_{5E}$        & $8.3\hspace{2pt}(7.8)$   & -            & $7.2\hspace{2pt}(6.5)$   & $6.2\hspace{2pt}(2.6)$    & $9.1\hspace{2pt}(5.4)$   & $3.2\hspace{2pt}(2.7)$    & $7.7\hspace{2pt}(3.1)$    & $5.3\hspace{2pt}(1.8)$    \\
$F_{E}$ [N]          & $1.9\hspace{2pt}(0.9)$    & -            & $1.9\hspace{2pt}(0.7)$   & -                         & $1.8\hspace{2pt}(0.4)$   & -                         & $2.6\hspace{2pt}(0.7)$    & -                         \\
$P_{E}$ [mm]         & $10.3\hspace{2pt}(2.3)$   & -            & $9.3\hspace{2pt}(1.1)$   & $9.9\hspace{2pt}(2.3)$    & $10.4\hspace{2pt}(4.9)$  & $10.2\hspace{2pt}(3.2)$   & $8.8\hspace{2pt}(0.9)$    & $8.1\hspace{2pt}(1.6)$    \\
$\tau_{1}$ [\%]      & $102.1\hspace{2pt}(68.8)$ & -            & $99.4\hspace{2pt}(1.1)^\dagger$  & $98.6\hspace{2pt}(2.8)^\dagger$   & $80.9\hspace{2pt}(22.2)^\dagger$ & $93.6\hspace{2pt}(12.4)^\dagger$   & $-52.7\hspace{2pt}(30.3)$ & $-56.7\hspace{2pt}(15.2)$ \\
$\tau_{2}$ [\%]      & $-44.6\hspace{2pt}(17.2)$ & -            & $-51.3\hspace{2pt}(4.1)$ & $-51.8\hspace{2pt}(4.2)$  & $-46.2\hspace{2pt}(4.8)$ & $-48.5\hspace{2pt}(5.5)$  & $-31.4\hspace{2pt}(1.7)$  & $-31.2\hspace{2pt}(3.4)$  \\
$\tau_{5}$ [\%]      & $-24.8\hspace{2pt}(10.4)$ & -            & $6\hspace{2pt}(1.4)$     & $21.1\hspace{2pt}(16.4)$  & $15.8\hspace{2pt}(13.6)$ & $22.8\hspace{2pt}(14.8)$  & $18.2\hspace{2pt}(21.4)$  & $17.9\hspace{2pt}(8.1)$   \\
BB                   & $5.86\hspace{2pt}(4.7)$   & -            & $6.52\hspace{2pt}(4.14)$ & $10.16\hspace{2pt}(5.89)$ & $6.81\hspace{2pt}(3.89)$ & $10.24\hspace{2pt}(6.31)$ & $4.14\hspace{2pt}(2.12)$  & $7.96\hspace{2pt}(5.04)$  \\
AD                   & $2.9\hspace{2pt}(2.8)$    & -            & $21.3\hspace{2pt}(8.6)$  & $20.5\hspace{2pt}(7.6)$   & $21.8\hspace{2pt}(9.6)$  & $24.2\hspace{2pt}(12.6)$  & $14.7\hspace{2pt}(8.6)$   & $7.5\hspace{2pt}(0.9)$    \\
MD                   & $3.3\hspace{2pt}(1.8)$    & -            & $20.2\hspace{2pt}(9.73)$ & $17.58\hspace{2pt}(5.73)$ & $14.23\hspace{2pt}(10)$  & $11.5\hspace{2pt}(6.12)$  & $12.6\hspace{2pt}(6.6)$   & $15.6\hspace{2pt}(8.01)$  \\
PD                   & $6.45\hspace{2pt}(4.54)$  & -            & $6.1\hspace{2pt}(2.98)$  & $5.35\hspace{2pt}(1.65)$  & $4.4\hspace{2pt}(2.75)$  & $5.12\hspace{2pt}(3.52)$  & $4\hspace{2pt}(2.3)$      & $3.9\hspace{2pt}(1.47)$   \\ \bottomrule
\end{tabular}
\end{center}
\vspace{-2.0mm}
\begin{tablenotes}
\item{$\theta_{1E}$, $\theta_{2E}$, and $\theta_{5E}$ are the error between optimized and the actual pose. $F_{E}$ is the force tracking error of the exoskeleton while $P_{E}$ is the RMS value of the cartesian error in $x,y,z$ axis of the robot. $\tau_{i}$ [\%] for $i=1,2,5$ represents the change of the torque at the defined joint.}
\vspace{-5.2mm}
\end{tablenotes}
\end{table*}

Next, in experiment-B, since $\bm{\tau_{Max}}$ is surpassed due to the change of payload (see Fig.\ref{fig:Fig.6} state A), the optimization provides a new arm configuration by extending the shoulder and elbow to distribute the torque load among those joints. Note that at state-B, joint torques, and EMG signals bear close and even smaller values than that of experiment-A within the Cartesian limits ($\bm{p}$) thanks to the control framework. The muscular activation also points out the aforementioned shoulder extension movement in Table.II, that is, the AD contracts $53.7\%$ less than that of PD.  Finally, the small increment of $\tau_{1}$ in Fig.\ref{fig:Fig.6} is due to the structure of the exoskeleton, which partially restricts the user from moving the arm to the fully adducted shoulder. This situation can be seen in its std values in Table.II as well, demonstrating a large fluctuation in the overall data.

Regarding the experiment-C results in Fig.\ref{fig:Fig.8}B and \ref{fig:Fig.5}, even though there is a sharp torque reduction about $\theta_{2}$, the optimized pose amplifies the torque about $\theta_{1}$ due to the assigned zero $W_{1}$. Therefore, MD is $82.4\%$ higher than that of experiment-B in Table.II. On the other hand, even though the values of $\tau_{5}$ are similar in both figures, the BB contraction is three times smaller in W/ assistance than W/O, showing the impact of the exoskeleton. Also, the small differences in robot Cartesian movements between both figures verifies the implementation of the same initial conditions (i.e., $\bm{\theta_{i}}$, and $\mathbb{W}$). 

When it comes to experiment-D results, $\tau_{1}$ decreases significantly (i.e., $5$ times smaller than experiment-C) because of nonzero $W_{1}$ as in both Fig.\ref{fig:Fig.8}A and \ref{fig:Fig.7}. Hence, MD is $31.8\%$ less than that of experiment-C according to Table.II, validating the abovementioned torque alleviation.

Afterward, according to experiment-E results in Fig.\ref{fig:Fig.9}, $\tau_{1}$ at state-A is greater than all the other experiments due to the initial nonzero $\theta_{1}$ values. Eventually, at state-B, the efforts in the shoulder and elbow are mitigated thanks to the developed control framework. Furthermore, the average $\tau_{1}$ reduction is notable, reaching $54.7\%$ ($\pm10.6$) in Table. II.

Finally, some general conclusions are drawn to demonstrate the effectiveness of the proposed control framework. For instance, in Table.II, the mean value ($\pm$ std) of $\theta_{1E}$, $\theta_{2E}$, $\theta_{5E}$, $F_{E}$, and $P_{E}$ of all experiments of participants are $5.22^{\circ}$ ($\pm 3.78$), $8.13^{\circ}$ ($\pm4.51$), $6.72^{\circ}$ ($\pm4.62$), $2.08$ N ($\pm0.37$ N), and $9.6$ mm ($\pm0.87$ mm), respectively. It is clear that the participants are guided via the avatar in the predetermined $\theta_{E}$ tolerance for three angles. Moreover, the small error of $F_{E}$ shows that the exoskeleton assistance is provided successfully to the subjects, and this error can be improved through more precise gain tuning. Furthermore, the marginal error of $P_{E}$ verifies the hand tracking performance of the robot. In addition, since a nonzero $W_{2}$ is defined for all tests, the average $\tau_{2}$ reduction and its standart deviation are substantial, bearing $43.6\%$ ($\pm5.8$). Another point is related to muscle activation. To clarify, as the initial $\theta_{2}$ values of experiment-C, and D are slightly different, AD contraction of them are close to each other (see Table.II). Yet, the initial $\theta_{2}$ of experiment-E is smaller than those experiments, and thus its AD contraction is $46.8\%$ less than that of those experiments. Finally, a major effort reduction is observed in BB muscle thanks to the elbow exoskeleton between W/ and W/O assistance, pointing $35.8\%$, $33.5\%$, and $47.9\%$ with respect to W/O for test-C, D and E, respectively. Furthermore, the rate of change of $\tau_{5}$ increases in most of the tests due to assigned zero $W_{5}$ except test-B. In this test, even though a nonzero $W_{1}$, and $W_{2}$ are set, since $\theta_{1}$ is already $0$ in the beginning, the main priority is given to the minimization of $\theta_{2}$. In this case, the optimization provides an arm configuration such that the center of mass of the upper and lower arm are almost aligned along the $z$ axis of the $\Sigma_{G}$ (see the arm configuration at state-B in Fig.\ref{fig:Fig.6}B ). To achieve this, the $\theta_{5}$ is reduced by the optimization, and hence $\tau_{5}$ is mitigated in test-B.


\vspace{-5.0mm}
\section{Conclusions \& Future Study}
\label{sec:sec.4}
\vspace{-1mm}
In this paper, we designed a control framework to alleviate the overloading joint torques in manual operations with the help of a cobot and a lightweight elbow exoskeleton. Through an anticipatory optimisation of the human arm configuration and cobot path planning, our framework enabled the users to align the external loads with the exoskeleton-supported joints to the maximum extent possible. Results show that the mean value of $\theta_{1E}$, $\theta_{2E}$, and $\theta_{5E}$ of all subjects of all experiments is $6.69\pm4.43^{\circ}$, concluding that the participants followed the visual feedback with minimal error. As a result of this small error, the mean value of the optimized joint torques of all subjects for $\tau_{1}$, $\tau_{2}$, and $\tau_{5}$ are $3.63{\pm} 0.85$ Nm, $6.83{\pm} 0.47$ Nm, and $7.2{\pm} 0.28$ Nm, respectively. It is clear that $\tau_{1}$ is slightly above the predetermined max threshold, yet $\tau_{2}$, and $\tau_{5}$ are significantly below than that of the $\tau_{Max}$ values. Given the support of the exoskeleton, which is $\approx6$ Nm for all subjects, the measured $\tau_{5}$ on participants is quite less, settling around $1$ Nm. The marginal increment of $\tau_{1}$ is due to the assigned $W_{1}=0$ in test-C, which increases the mean value. However, we assigned it intentionally to demonstrate the performance of the optimization under different $\mathbb{W}$. 

Regarding the limitations of our framework, it cannot be applied in a task where a human bends or takes a step to move to another position due to the predetermined calibration location. Thus, this calibration process should be improved so that the  method can be implemented in a variety of applications in real workstations. Also, we developed the human model for quasi-static movements, neglecting the dynamic effects. Hence, future studies will overcome those challenges by extending the framework to multi DoFs (full-body model) with several lightweight exoskeletons while varying $\mathbb{W}$ adaptively in a dynamic use case, and performing statistical analysis (e.g., NASA) on a higher number of subjects.

\vspace{-4mm}
\bibliographystyle{IEEEtran}
\bibliography{RAL_submitted}

\end{document}